\documentclass[conference]{IEEEtran}
\IEEEoverridecommandlockouts

\usepackage{cite}
\usepackage{amsmath,amssymb,amsfonts}
\usepackage{algorithmic}
\usepackage{graphicx}
\usepackage{textcomp}
\usepackage{xcolor}
\usepackage{bm}
\usepackage{multirow}
\usepackage{tabularx, colortbl, makecell}
\usepackage{etoolbox}
\usepackage{hyperref}

\usepackage{pifont}
\newcommand{\cmark}{\ding{51}}%
\newcommand{\xmark}{\ding{55}}%

\usepackage[font=small,labelfont=bf]{caption}

\def\BibTeX{{\rm B\kern-.05em{\sc i\kern-.025em b}\kern-.08em
    T\kern-.1667em\lower.7ex\hbox{E}\kern-.125emX}}
\begin{document}

\title{LauraTSE: Target Speaker Extraction using Auto-Regressive Decoder-Only Language Models
\thanks{Ming Li is the corresponding author. This research is funded by DKU foundation project "Emerging AI Technologies for Natural Language Processing". Many thanks for the computational resource provided by the Advanced Computing East China Sub-Center.}
}

\author{


\IEEEauthorblockN{Beilong Tang\IEEEauthorrefmark{1}, Bang Zeng\IEEEauthorrefmark{1}, Ming Li\IEEEauthorrefmark{1}}
\IEEEauthorblockA{\IEEEauthorrefmark{1}Duke Kunshan University, Kunshan, China \\
\{beilong.tang, zeng.bang, ming.li369\}@dukekunshan.edu.cn}
\vspace{-10pt}

}

\maketitle

\begin{abstract}
We propose LauraTSE, an Auto-Regressive Decoder-Only Language Model for Target Speaker Extraction built upon the LauraGPT backbone. LauraTSE employs a small-scale auto-regressive decoder-only language model that generates the initial layers of the target speech's discrete codec representations from the continuous embeddings of both the mixture and reference speech. These outputs serve as coarse-grained predictions. To refine them, a one-step encoder-only language model reconstructs the full codec representation by integrating information from both the mixture and the reference speech, adding fine-grained details. Experimental results show that our approach can achieve promising performance. Additionally, we conduct ablation studies to investigate the data scalability and the contribution of the encoder-only model.

\end{abstract}


\section{Introduction}
Target Speaker Extraction (TSE) aims at extracting target speaker's speech from a mixture using auxiliary information like reference speech, spatial information, or visual information etc., regarding the target speaker \cite{zmolikova2023neural}. Current dominant approaches utilize discriminative models which try to directly map the mixture speech to target clean speech \cite{luo2019conv,spex_plus,sepformer,sef_net}. However, this method might struggle for unseen data and sometimes even introduce undesirable distortions \cite{distortion}. Also, when data is highly corrupted with a low Signal-to-Noise Ratio (SNR), directly mapping might not be optimal. Generative models, on the other hand, have gained the attention for its capability in dealing with unseen noises compared with discriminative models \cite{fang2021variational, bie2022unsupervised, lu2022conditional} as well as its superior performances in terms of the audio quality \cite{target_diff,tokensplit, zhang2025anyenhance,kang2025llase}. Rather than learning the map from noisy speech to clean speech, generative models aim at learning the underlying distribution of the clean output. Generative models like diffusion models \cite{target_diff}, variational autoencoders (VAEs) \cite{vae}, and language models (LMs) \cite{tang2024tselm} have been studied for TSE. TSELM \cite{tang2024tselm} utilizes discrete tokens from WavLM \cite{wavlm} and encoder-only LMs for TSE \cite{tang2024tselm}. AnyEnhance \cite{zhang2025anyenhance} utilizes a masking encoding LM for multi-task speech processing, including TSE. 

However, Auto-Regressive (AR) decoder-only LMs, as another important class of generative models, have not been well-studied for TSE. One of the existing works related is SpeechX \cite{wang2024speechx}, which proposes a multi-task speech processing model utilizing an AR decoder-only LM. However, this work may still have several limitations. Firstly, the TSE task in the paper remains relatively simple \cite{wang2024speechx}, which does not demonstrate the full capability of AR decoder-only LMs on TSE tasks. Secondly, this work uses discrete representations as the input to the AR decoder-only LM, which turns out to be suboptimal for certain tasks compared with continuous features as mentioned in \cite{du2023lauragpt}. Finally, since SpeechX is designed as a multi-task system—handling tasks such as noise suppression, speech removal, TSE, and TTS—it remains unclear whether a small-scale single-task AR decoder-only LM can effectively perform TSE on its own.

To use AR LMs, we need to discretize audio into tokens for the classification loss. There are two main approaches for audio discretization. The first approach applies Kmeans clustering on the outputs of self-supervised learning (SSL) models, as done in \cite{selm, tang2024tselm, unit_hifi, dasb}. However, this method has been shown to lose speaker-specific information \cite{tang2024tselm, dasb}, likely because Kmeans discretization discards fine-grained speaker details while retaining mainly semantic information. The second approach leverages the Residual Vector Quantization (RVQ) layers of the neural audio codec \cite{du2024funcodec, wang2024speechx, du2023lauragpt}, which discretizes audio into multiple layers of finite token sequences. This method has shown greater promise in preserving both acoustic and speaker-related information \cite{du2023lauragpt}, making it a potential for tasks like TSE where the speaker information needs to be well-preserved.

\begin{figure*}[t]
  \centering
  \includegraphics[width=0.8\textwidth]{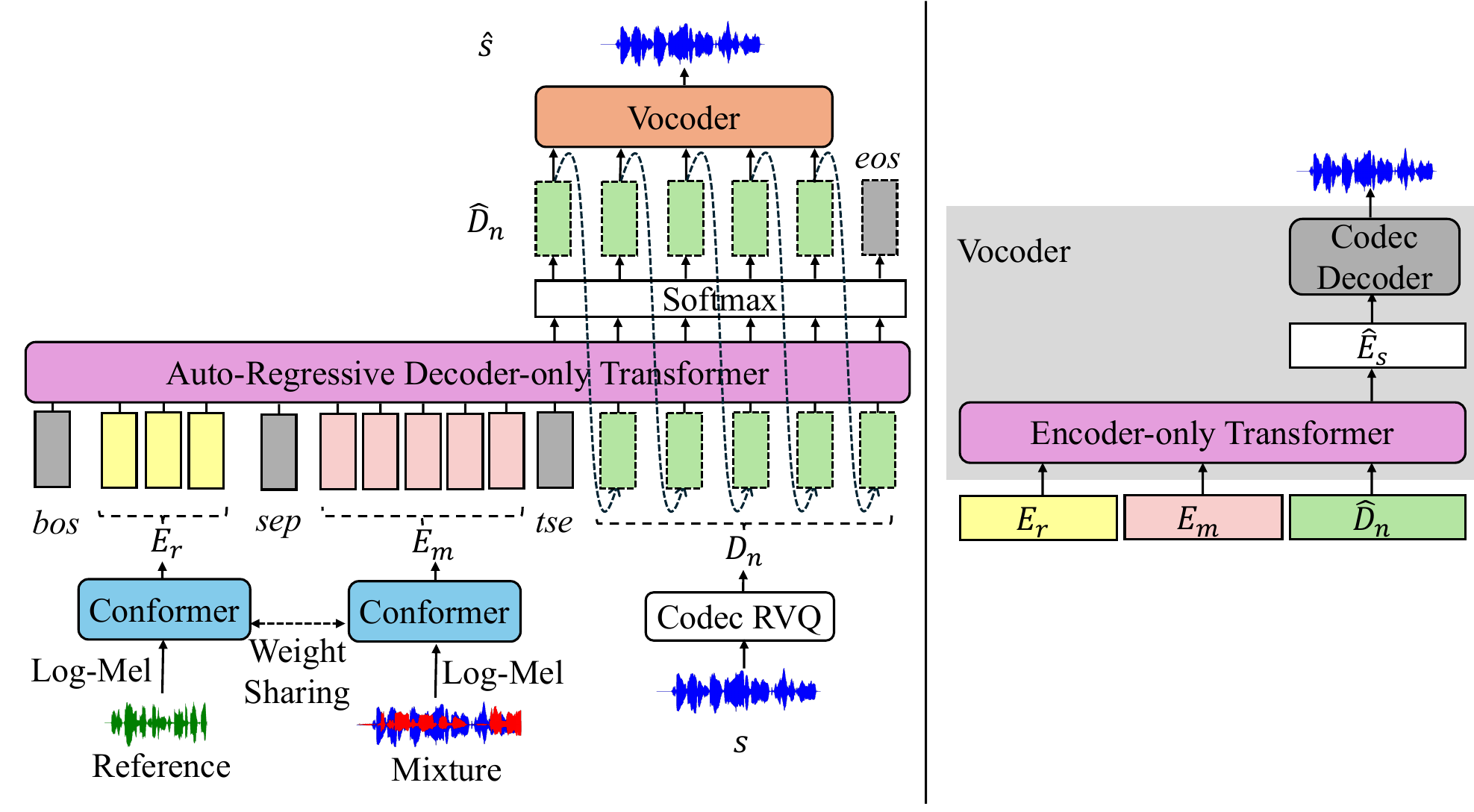}
  \caption{Overview of LauraTSE.}
  \label{model}
  \vspace{-5pt}
  \end{figure*}

In this work, we propose LauraTSE, an AR decoder-only LM for TSE, built upon the LauraGPT backbone \cite{du2023lauragpt}. The model takes the log-mel spectrograms of both the reference and the mixture speech as input, and utilizes a neural audio codec to represent the output of the AR model. LauraTSE consists of two components: (1) an AR decoder-only LM that predicts the first few layers of the codec representations of the target speech, and (2) a one-step encoder-only LM that directly predicts the sum of all layers of the codec embeddings by integrating information from both the mixture and the reference signals. An overview of the model architecture is provided in Figure~\ref{model}. To the best of our knowledge, we are the first to conduct single-task TSE using AR decoder-only LMs with continuous input features. Experimental results show that LauraTSE achieves promising performance in speech quality, speech intelligibility and speaker similarity. Our demos and code are available at \footnote{\href{https://beilong-tang.github.io/lauraTSE.demo/}{https://beilong-tang.github.io/lauraTSE.demo/}}.

\section{Method}

\subsection{Encoder}

The first stage of LauraTSE is the encoding stage. Similar to the Speech Enhancement (SE) tasks in LauraGPT \cite{du2023lauragpt}, we begin by computing the log-mel spectrograms for both the reference and mixture speech signals. These spectrograms are then passed through a shared Conformer \cite{gulati2020conformer} model, which produces continuous embeddings for the reference and mixture, denoted as 
$E_r$ and $E_m$, respectively. This stage acts as an adapter, transforming both the mixture and the reference inputs into a suitable representation space for the subsequent AR decoder-only LM.
Notably, unlike SpeechX \cite{wang2024speechx}, which utilizes discrete embeddings obtained from a neural audio codec, our approach follows \cite{du2023lauragpt} which retains continuous representations learned directly from the task. Our experimental results show that using continuous features as inputs leads to better performance compared to discrete token embeddings.

\subsection{Auto-Regressive Decoder-Only Language Model}

\begin{table*}
  \caption{Results on Libri2Mix clean.  In the "Category" column, "G" refers to generative models, while "D" refers to discriminative models.}
  \renewcommand{\arraystretch}{1.2}
  \begin{center}
    \setlength{\tabcolsep}{7.5pt}
    \begin{tabular}{ccccccccccc}
      \Xhline{2\arrayrulewidth}
      \multirow{2}{*}{Model} & \multirow{2}{*}{Category} & \multicolumn{3}{c}{DNSMOS $\uparrow$} & \multirow{2}{*}{NISQA $\uparrow$} & \multirow{2}{*}{SpeechBERT $\uparrow$} & \multirow{2}{*}{dWER $\downarrow$} & \multirow{2}{*}{WavLM Sim $\uparrow$} & \multirow{2}{*}{Wespeaker Sim $\uparrow$} \\
      \cline{3-5}
                             &                           & SIG & BAK & OVL &                         &                      &                         &                          &                          \\
      \hline
      Mixture                & -                         & 3.383 & 3.098 & 2.653 & 2.453 & 0.572 & 0.792 & 0.847 & 0.759 \\
      \hline
      Spex+\cite{spex_plus}  & D                         & 3.472  & 4.027  & 3.186  & 3.349 & 0.878 & 0.148 & 0.973 & 0.935 \\
      WeSep\cite{wang2024wesep} & D                      & 3.486 & 3.838 & 3.118 & 3.892 & 0.895 & 0.123     & 0.980 & 0.945 \\
      USEF-TSE \cite{usef_tse} & D                      & 3.555 & 4.051 & 3.272 & 4.319 & \textbf{0.935} & \textbf{0.0747}     & \textbf{0.988} & \textbf{0.968} \\
      \hline
      TSELM-L \cite{tang2024tselm} & G                   & 3.489 & 4.041 & 3.212 & 3.961 & 0.793 & 0.297 & 0.887 & 0.627 \\
      AnyEnhance \cite{zhang2025anyenhance} & G          & \textbf{3.638} & 4.066 & \textbf{3.353} & 4.277 & 0.735 & -     & 0.914 & - \\
      \hline
      LauraTSE               & G                         & 3.609 & \textbf{4.084} & 3.336 & \textbf{4.333} & 0.908 & 0.159 & 0.974 & 0.876 \\
      LauraTSE-streaming & G & 3.596 & 4.061 & 3.314 & 4.275 & 0.897 & 0.169 & 0.973 & 0.873 \\
      \Xhline{2\arrayrulewidth}
    \end{tabular}
    \label{tbl:libri2mix_main_exp}
  \end{center}
  \vspace{-10pt}
\end{table*}

Our AR decoder-only LM aims to predict the joint probability distribution of the coarse discrete representation of the target speech $s$ conditioned on the reference 
speech and the mixture speech embedding 
according to the probability chain rule:
$$
P_\theta(\hat{D}_n\mid E_r,E_m) = \prod_{i\le T}{P_\theta(\hat{D}_n^{(i)}\mid \hat{D}_n^{(1:i-1)}, E_r, E_m)}
$$
where $T$ denotes the length of the output signal, and $\theta$ denotes the model parameters, and $\hat{D}_n$ denotes the generated discrete representation of the target speech.

During training, the input sequence to the AR decoder-only LM is formatted as $ [\textit{bos} , E_r, \textit{sep}, E_m, \textit{tse}, D_{n}] $, where $bos$ is a learnable token representing the start of the sentence, and $sep$ is a token that separates the reference embeddings and the mixture embeddings. $tse$ is used to split the given input and the generated output. $D_n$ is the sum of the embeddings of the first $n$ layers of the RVQ output embeddings of the target speech.  The AR model is trained to generate the first 
$n$ layers of the discrete tokens of the target speech. More specifically, after the decoder-only Transformer generates the output embeddings, we utilize $n$ linear layers to get the distribution of the first $n$ layers of the RVQ of the target speech. 

The training objective is the Cross-Entropy loss between the predicted tokens and the first $n$ layers of the ground-truth discrete tokens from the RVQ. Once these $n$ layers of discrete tokens are predicted, they are mapped to their corresponding embeddings via the neural audio codec's own embedding table. These $n$ layers of discrete embeddings are then summed to one single-layer embedding $\hat{D}_n$ as the output. During inference, the AR decoder-only LM generates $\hat{D}_n$ frame by frame. 

By choosing $n$ to be smaller than the total number of RVQ layers, we simplify the modeling task, allowing the AR decoder to focus on generating coarse-grained representations of the target speech while still preserving intelligibility.

\subsection{Vocoder}

The goal of the vocoder is to reconstruct the clean audio waveform from the coarse representations generated by the AR model, utilizing both the mixture and reference speech embeddings. The vocoder consists of an encoder-only LM and a frozen pretrained codec decoder. The encoder-only LM employs the self-attention mechanism to capture fine-grained acoustic details. Following \cite{du2023lauragpt}, we adopt a one-step encoder LM to directly predict the sum of all the RVQ layers of the target speech's codec embedding, instead of generating them layer-by-layer as done in \cite{wang2024speechx}.  Specifically, it takes the concatenated input $[E_r, E_m, \hat{D}_n]$—representing reference, mixture, and generated coarse embeddings—and outputs $[., ., \hat{E}_s]$, where $\hat{E}_s$ is the predicted fine-grained embeddings of the target speech. We apply both the L1 and L2 loss between $\hat{E}_s$ and the ground-truth embedding $E_s$ of the target speech from the sum of all the RVQ layers. Finally, the pretrained codec decoder converts the predicted embeddings into the target raw waveform. Note that the decoder-only LM and the encoder-only LM are jointly optimized during training.

\subsection{Streaming Inference}

\begin{figure}[t]
\centering
\includegraphics[width=0.4\textwidth]{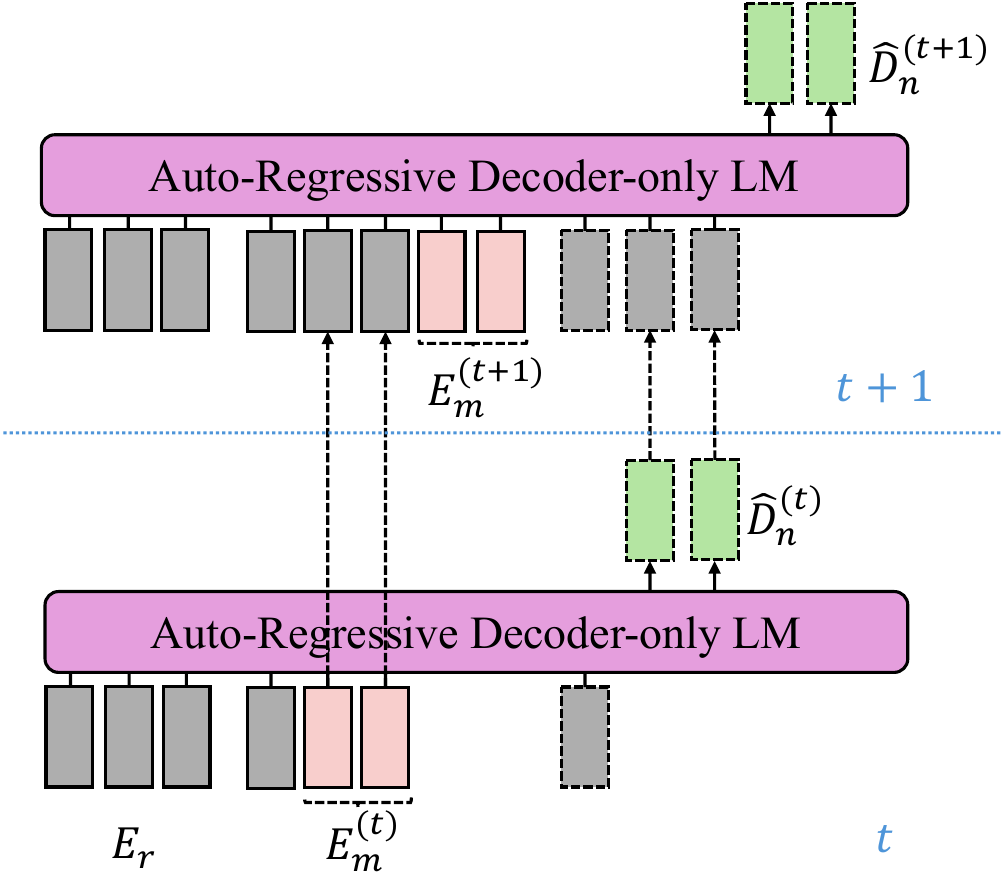}
\caption{Streaming inference procedure of LauraTSE (\textit{LauraTSE-streaming}). Pink and green highlights denote the current input mixture representation $E_m^{(t)}$ and output representation $\hat{D}_n^{(t)}$ at timestep $t$, respectively}
\label{fig:laura_causal}
\vspace{-10pt}
\end{figure}

Real-time TSE is crucial for practical applications \cite{sato2024speakerbeam}. Since AR decoder-only LM can be easily adopted for streaming inference \cite{tsunoo2024decoder}, we adopt LauraTSE and propose \textit{LauraTSE-streaming}, which processes the input mixture speech $s_m$ incrementally in $N$ sequential chunks $\{s_m^{(1)}, s_m^{(2)}, \dots, s_m^{(N)}\}$. As illustrated in Figure~\ref{fig:laura_causal}, at each timestep $t$:
\begin{enumerate}
    \item Receive new causal chunk $s_m^{(t)}$
    \item Encode to obtain $E_m^{(t)}$
    \item Concatenate $E_m^{(t)}$ with previous representations to get $E_m^{(1:t)}$
    \item Use the decoder-only LM's output from all the previous timesteps as the prompt
    \item Generate current output representation $\hat{D}_n^{(t)}$
    \item Utilize the Vocoder to reconstruct the raw waveform of $\hat{D}_n^{(t)}$
\end{enumerate}
This iterative process continues until all chunks are processed, enabling frame-by-frame streaming.

\section{Experiments setup}

\subsection{Dataset}

Our main experiments are conducted on the 460-hour clean speech subset of LibriSpeech \cite{librispeech} (LibriSpeech-460h). The training data is generated on-the-fly by mixing speech samples with a relative SNR randomly sampled between 0 and 5 dB. For cross-validation, we use the clean dev set from Libri2Mix \cite{librimix}. During both training and evaluation, the reference audio is randomly clipped to 5 seconds. For evaluation, we use the clean test set of Libri2Mix. Reference utterances are randomly selected for each target speaker to simulate realistic target speaker extraction conditions. Note that LauraTSE is firstly trained on LibriSpeech-460h and then finetuned on Libri2Mix clean training set. For ablation studies, our training is conducted on the Libri2Mix clean training set. 

\subsection{Model Details}

We adopt LauraGPT \cite{du2023lauragpt} as the backbone for our AR decoder-only LM, and use FunCodec \cite{du2024funcodec} as the neural audio codec. We set the AR output layer number $n$ as 2. For encoding, we apply a hop size of 256 and a window size of 512 to both the reference and mixture speech. The conformer encoder consists of 6 layers, each with 8 attention heads and a hidden dimension of 512. The decoder-only transformer has 10 layers, 8 attention heads, and a hidden size of 512. The encoder-only transformer used in the vocoder also comprises 6 layers with 8 attention heads and a 512-dimensional feature space. We set the chunk size of LauraTSE-streaming as 2 seconds. 

Our LauraTSE model is trained from scratch. It contains a total of 77M parameters, with 36M allocated to the decoder-only transformer. We use the Adam optimizer with an initial learning rate of $1 \times 10^{-3}$. A warm-up scheduler with 10,000 warm-up steps is employed, and the learning rate is halved if the evaluation performance does not improve within 3 consecutive epochs. Training is conducted on 16 GPUs, each equipped with 32GB of memory, for a total of 100 epochs.

\subsection{Evaluation Metrics}
Traditional metrics such as PESQ \cite{rix2001perceptual}, SI-SNR, and STOI \cite{taal2010short} are not used due to the potential misalignment between the generated waveform from vocoders and the original waveform \cite{selm}. Therefore, we use:

\begin{itemize}
\item \textbf{DNSMOS} \cite{dnsmos}: a reference-free metrics that has three scores from 1 to 5: SIG, BAK, and OVL, representing the signal quality, background noise and the overall quality, respectively.
\item \textbf{NISQA} \cite{mittag2021nisqa}: Another reference-free metric that predicts an overall quality score between 1 and 5 for the generated speech.
\item \textbf{SpeechBERT} \cite{saeki2024speechbertscore}: A semantic similarity metric based on BERTScore computed over self-supervised speech representations. We use the HuBERT-base \cite{hubert} model to extract the features for comparison between the generated and the target speech.
\item \textbf{Differential Word Error Rate (dWER)} \cite{dwer}: This metric computes the word error rate between the generated and the ground-truth speech using an ASR model. We employ the \textit{base} model of Whisper \cite{whisper} for this evaluation.
\item \textbf{Speaker Similarity}: This metric measures the speaker similarity between the output and ground-truth speech via cosine similarity over high-dimensional embeddings. We use two models for this task: the WavLM-base\footnote{\href{https://huggingface.co/microsoft/wavlm-base-plus-sv}{https://huggingface.co/microsoft/wavlm-base-plus-sv}} and the \textit{Resnet\_221LM} model from WeSpeaker \cite{wespeaker}.

\end{itemize}

\subsection{Baseline models}
We compare LauraTSE with several recent baselines. We include Spex+ \cite{spex_plus}, a classic discriminative TSE model. We also compare LauraTSE with two recent discriminative models: USEF-TSE \cite{usef_tse} and the BSRNN model from WeSep \cite{wang2024wesep}. Additionally, we compare it with two generative models: TSELM-L \cite{tang2024tselm} and AnyEnhance \cite{zhang2025anyenhance}. AnyEnhance is a multi-task model based on masked generative modeling. Note that Spex+, TSELM-L, and USEF-TSE are trained on LibriSpeech-460h. For WeSep, we directly use the public checkpoint trained on Voxceleb \cite{voxceleb}. For AnyEnhance, we directly use the TSE results provided in their work. 

\section {Results and Discussion}

\subsection{Comparison with other baseline models}

Table~\ref{tbl:libri2mix_main_exp} presents the overall results. LauraTSE achieves comparable speech quality to AnyEnhance, while outperforming it in both the speaker similarity (WavLM Sim) and the semantic similarity (SpeechBERT). Notably, LauraTSE is trained on only 460 hours of data, whereas AnyEnhance is trained on 5000 hours, raising the question of whether multi-task learning is always superior to task-specific models for a particular objective. Compared with TSELM-L \cite{tang2024tselm}, which uses discrete tokens from WavLM, our model leverages the neural audio codec representation that better preserves the speaker identity, addressing the issue of low speaker similarity. Additionally, LauraTSE outperforms discriminative baselines such as USEF-TSE \cite{usef_tse} and WeSep \cite{wang2024wesep} in terms of speech quality, while maintaining competitive performance in semantic similarity and speaker consistency. However, there is a notable gap of the semantic similarity (dWER) between LauraTSE and USEF-TSE.

\subsection {Data Scalability of LauraTSE}

\begin{figure}
\centering
\caption{dWER versus training data scale across models. Annotations "(-X)" denote relative dWER reduction (percentage points) compared to the preceding smaller dataset.}
\includegraphics[width=0.40\textwidth]{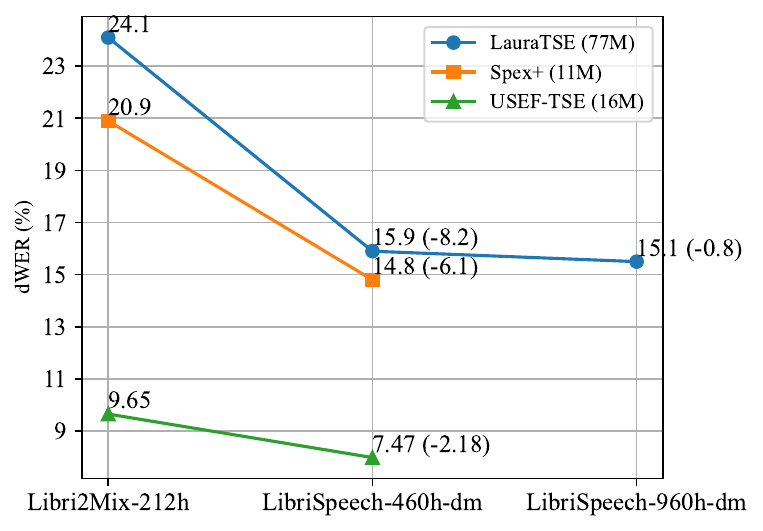}
\label{fig:data_scale}
\vspace{-10pt}
\end{figure}

AR generative models have demonstrated strong scalability with respect to training data size\cite{data_scale_prove}. To evaluate this, we compare the data scalability of LauraTSE against two representative discriminative models: Spex+ and USEF-TSE, which contains 11M and 16M parameters, respectively. 

We train LauraTSE on three datasets of increasing size: \textit{Libri2Mix-212h}, \textit{LibriSpeech-460h-dm}, and \textit{LibriSpeech-960h-dm}. Specifically, \textit{Libri2Mix-212h} uses the Libri2Mix-clean subset, containing approximately 212 hours of audio. For \textit{LibriSpeech-460h-dm}, we use the LibriSpeech-460h subset with dynamic mixing. Similarly, \textit{LibriSpeech-960h-dm} includes the full 960-hour LibriSpeech dataset, also with dynamic mixing applied.


Results (Figure~\ref{fig:data_scale}) indicate that while the discriminative model USEF-TSE \cite{usef_tse} exhibits relatively stable performance as the training set scales from Libri2Mix to LibriSpeech, both Spex+ \cite{spex_plus} and LauraTSE show significant gains. LauraTSE achieves an 8.2\% ($24.1-15.9=8.2$) absolute dWER reduction—surpassing the 6.1\% improvement of Spex+—highlighting its superior scalability with larger datasets.

It is also worth noticing that the performance gains by increasing the training data from 460 hours to 960 hours are relatively modest. One possible explanation is that the LibriSpeech-500-other subset, which constitutes a significant portion of the 960-hour dataset, is less clean and more acoustically diverse than the LibriSpeech-460h subset, potentially introducing noise that limits model improvement. Another contributing factor could be that the current model size is insufficient to fully leverage the additional data. To better understand how data scalability affects the model performance, future research is needed. Moreover, LauraTSE does not achieve better performance compared with USEF-TSE even with more training data, raising the question of whether generative models can outperform discriminative models in semantic similarity. 

\begin{figure*}
\centering
\caption{Illustration of different input types for the encoder-only LM. 
\textit{Encoder-All} corresponds to the LauraTSE baseline, where the encoder-only LM receives both the mixture and reference embedding.
\textit{Encoder-Mix} uses only the mixture embedding, while \textit{Encoder-Ref} uses only the reference embedding.  }
\includegraphics[width=0.6\textwidth]{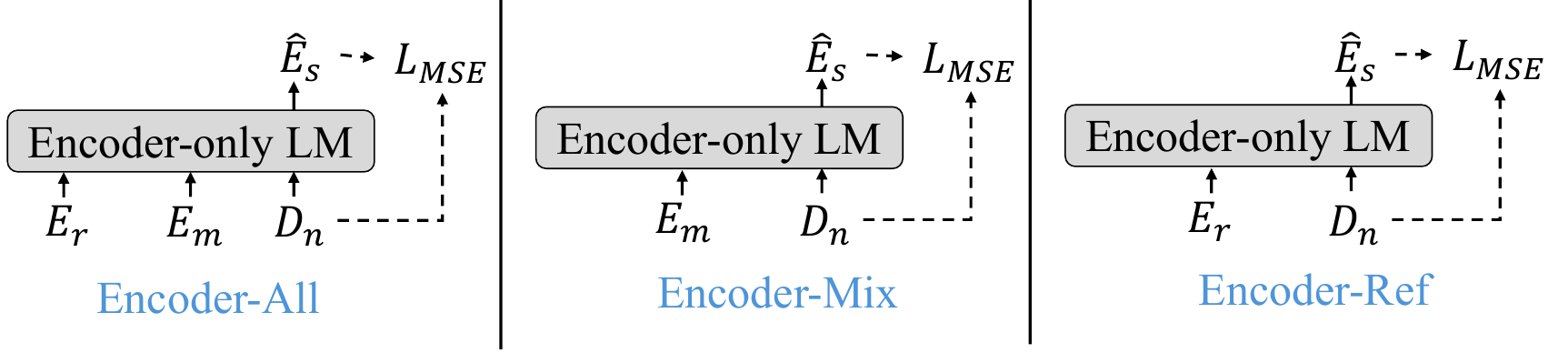}
\label{fig:encoder_input}
\vspace{-5pt}
\end{figure*}
\begin{figure}
\centering
\caption{Decoder-Encoder Joint vs. Split. \textit{Decoder-Encoder-join} denotes the proposed LauraTSE model where Decoder and Encoder are trained together by Cross-Entropy Loss and MSE Loss. In \textit{Decoder-Encoder-split}, the Decoder and Encoder is trained separately. }
\includegraphics[width=0.4\textwidth]{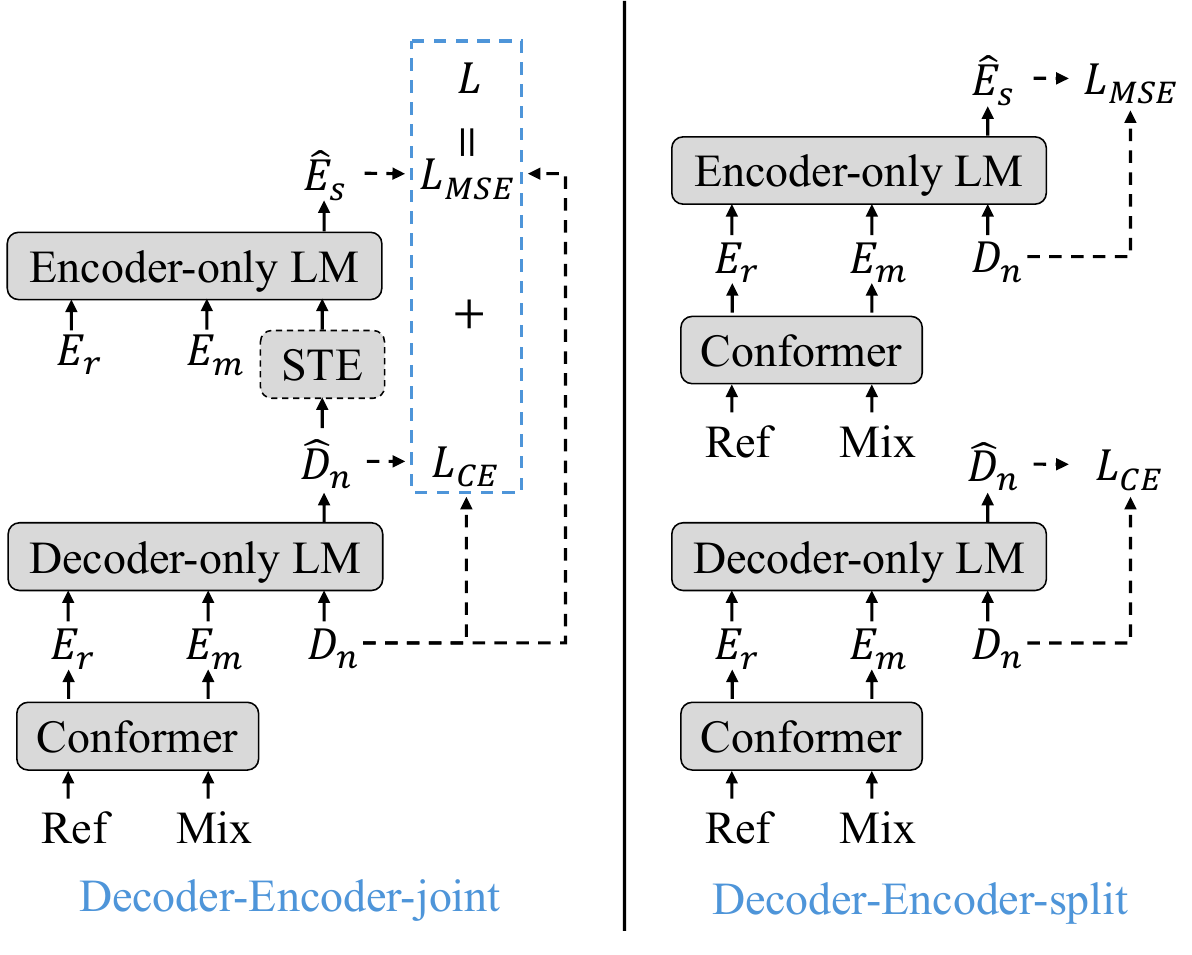}
\label{fig:encoder_decoder_split}
\vspace{-15pt}
\end{figure}

\begin{table}
  \renewcommand{\arraystretch}{1.2}
  \begin{center}
      \caption{
  Evaluation results for different Decoder-Encoder configurations. \textit{Decoder-Encoder-joint} and \textit{Decoder-Encoder-split} refer to the two integration strategies illustrated in Figure~\ref{fig:encoder_decoder_split}. \textit{Target-$n$} denotes the reconstructed target clean audio using only the first $n$ layers of the codec. \textit{No-Encoder} uses summation of only the first $n$ layers of the decoder-only LM output to generate speech without the encoder.
  }
    \begingroup
    \setlength{\tabcolsep}{6pt}  
    \begin{tabular}{cccc}
      \Xhline{2\arrayrulewidth}
      Model & NISQA $\uparrow$ & dWER $\downarrow$ & WeSpeaker Sim $\uparrow$ \\
      \hline
      Decoder-Encoder-joint    & 4.241 & 0.241 & 0.847 \\
      Decoder-Encoder-split    & 4.253 & 0.232  & 0.858 \\
      \hline
      Target-$n$ ($n=2$)              & 3.644 & 0.301 & 0.740 \\
      No-Encoder               & 3.807 & 0.579 & 0.709 \\
      \Xhline{2\arrayrulewidth}
    \end{tabular}
    \endgroup
    \label{tbl:encoder_split}
  \end{center}
\end{table}

\begin{table}
  \caption{Input composition results for the encoder-only LM.  For more information, please refer to Figure~\ref{fig:encoder_input}.}
  \renewcommand{\arraystretch}{1.2}
  \begin{center}
    \setlength{\tabcolsep}{5pt}
    \begin{tabular}{cccccccc}
      \Xhline{2\arrayrulewidth}
      \multirow{2}{*}{Model} & \multicolumn{3}{c}{Input} & \multirow{2}{*}{NISQA$\uparrow$} & \multirow{2}{*}{dWER $\downarrow$} & \multicolumn{1}{c}{WeSpeaker} \\
      & $E_r$ & $E_m$ & $D_n$ & & & Sim$\uparrow$ \\
      \hline
      Encoder-All   & \cmark & \cmark & \cmark & 4.241 & 0.241 & 0.847 \\
      Encoder-Mix      & \xmark & \cmark & \cmark   & 4.173 & 0.239 & 0.842 \\
      Encoder-Ref      & \cmark & \xmark & \cmark  & 4.187 & 0.480 & 0.763 \\
      \Xhline{2\arrayrulewidth}
    \end{tabular}
    \label{tbl:encoder_input}
  \end{center}
  \vspace{-10pt}
\end{table}

\begin{table*}
  \caption{Ablation studies of LauraTSE. \textit{n-} denotes the output layer number of the decoder-only LM. The \textit{Ref output} formats the output of the decoder-only LM to contain both the clean and reference speech. \textit{Discrete IO} uses discrete codec embeddings rather than continuous features as the input features. For \textit{WavLM input}, the WavLM \cite{wavlm} embeddings are utilized as the input features.}
  \renewcommand{\arraystretch}{1.2}
  \begin{center}
    \setlength{\tabcolsep}{8pt}
    \begin{tabular}{cccccccccc}
      \Xhline{2\arrayrulewidth}
      \multirow{2}{*}{Model} & \multicolumn{3}{c}{DNSMOS $\uparrow$} & \multirow{2}{*}{NISQA $\uparrow$} & \multirow{2}{*}{SpeechBERT $\uparrow$} & \multirow{2}{*}{dWER $\downarrow$} & \multirow{2}{*}{WavLM Sim $\uparrow$} & \multirow{2}{*}{Wespeaker Sim $\uparrow$} \\
      \cline{2-4}
                             & SIG & BAK & OVL &                         &                      &                         &                          &                          \\
      \hline
      Base ($n$-2)          & 3.626 & 4.102 & 3.360 & 4.241 & 0.880 & 0.241 & 0.965  & 0.847 \\
      $n$-1   & 3.604 & 4.100 & 3.339 & 4.201 & 0.861 & 0.266 & 0.958  & 0.830 \\
      $n$-3    & 3.618 & 4.095 & 3.350 & 4.270 & 0.880 & 0.235 & 0.967  & 0.853 \\
      \hline
      Ref output    & 3.588 & 4.071 & 3.318 & 4.182 & 0.859 & 0.237 & 0.962 & 0.851 \\
      Discrete IO & 3.562 & 4.035 & 3.268 & 3.940 & 0.810 & 0.421 & 0.952  & 0.835 \\
      WavLM input  & 3.507 & 3.951 & 3.137 & 3.220 & 0.792 & 0.447 & 0.860 & 0.633        \\
      \Xhline{2\arrayrulewidth}
    \end{tabular}
    \label{tbl:ablation}
  \end{center}
  \vspace{-10pt}
\end{table*}

\subsection{What role does the Encoder-only LM play?}

Our ultimate goal is to perform TSE using only a decoder-only LM, without relying on an additional encoder-only LM. However, it is difficult for the single-layer neural audio codec to preserve fine-grained speech details as well as multi-layer codecs. Therefore, we need to add the encoder-only LM to refine the coarse-grained output from the decoder-only LM into high-resolution, continuous embeddings. This section studies the effect of the encoder-only LM.

\subsubsection{Joint vs. Separate Training}

Firstly, we are interested in whether decoder-only LM and encoder-only LM needs to be trained jointly as their tasks are different. We conduct the following experiments: \textit{Decoder-Encoder-joint} and \textit{Decoder-Encoder-split}. Details can be found at Figure~\ref{fig:encoder_decoder_split}. \textit{Decoder-Encoder-joint} is the original LauraTSE model where the encoder-only LM and the decoder-only LM are trained jointly through a Straight-Through Estimator (STE) module \cite{ste} which ensures the gradient flow for softmax. For \textit{Decoder-Encoder-split}, the decoder-only LM and the encoder-only LM are trained separately. The results can be found at Table~\ref{tbl:encoder_split}. We find that training decoder-only LM and encoder-only LM separately achieves slightly superior performance compared with training them jointly, suggesting it not necessary to train them together. 

\subsubsection{Input composition for Encoder-only LM}

The current input to the encoder-only LM contains both the mixture and the reference. To fully study the role of the encoder-only LM, we need to study which input speech information is more important. Therefore, we compose three systems: \textit{Encoder-All}, \textit{Encoder-Mix}, and \textit{Encoder-Ref}. \textit{Encoder-All} is the original encoder that takes both the mixture and the reference. For \textit{Encoder-Mix}, only the mixture embedding is fed to the encoder, while \textit{Encoder-Ref} only takes the reference embedding. Details can be found at Figure~\ref{fig:encoder_input}. 

The results are shown at Table~\ref{tbl:encoder_input}. We can see that the \textit{Encoder-Mix} achieves similar performance as \textit{Encoder-All}, while \textit{Encoder-Ref} achieves much worse performance. This result indicates that the mixture embedding is crucial for the encoder-only LM, and the reference embedding might not be necessary here. 

\subsubsection{Does Encoder-only LM help the TSE?}

Given that the mixture representation for the encoder-only LM is crucial for the final results, we assume that the encoder-only LM not only generates the fine-grained representation from the coarse output of the decoder-only LM, but more importantly, it also aids the TSE task. To further validate this, we compare:
\begin{itemize}
    \item \textit{Target-n:} Clean target speech reconstructed using only the first $n$ layers of the codec. Same as the original LauraTSE, we set $n$ as 2.
    \item \textit{No-Encoder:} Speech reconstructed from the first $n$ layers output codec from the decoder-only LM only.
    \item \textit{Decoder-Encoder-split:} The encoder-only LM is applied after the prediction from the decoder-only LM.
\end{itemize}

As shown in Table~\ref{tbl:encoder_split}, \textit{No-Encoder} underperforms compared to \textit{Target-n}, indicating that the decoder-only LM alone is insufficient for this task. Adding back the encoder-only LM leads to a significant improvement (Decoder-Encoder-split), reducing the dWER from 0.579 to 0.232.

These findings suggest that the encoder-only LM is currently essential—not only for reconstructing fine-grained speech features but also for assisting with the TSE task. Our future works will focus on developing techniques that allow the decoder-only LM to perform TSE effectively without relying on an additional encoder.

\subsection{Other Ablation Studies}

Table \ref{tbl:ablation} shows the results of other ablation studies we conduct. \textit{Base} refers to the proposed LauraTSE model. \textit{$n$-} refers to the output number of layers of the AR decoder-only LM. Changing $n$ from 1 to 3 results in minimal performance differences, suggesting that even a small number of coarse layers may provide sufficient information for TSE.

AR decoder-only LMs like LauraGPT \cite{du2023lauragpt} conduct tasks like SE where the output is strictly aligned with the input length. To study if the output length has to be the same as the input, we have formatted the decoder-only input to be $[\textit{bos}, E_r,E_m,\textit{tse}]$.
Unlike the original method, which only produces the clean speech, this approach concatenates the reference and the mixture speech into a single input sequence, expecting the model to generate an output containing both the reference speech and the enhanced speech. This method is referred to as the \textit{Ref output}. During inference, we retain only the portion of the output corresponding to the mixture, constrained by the length of the reference speech. This approach yields results similar to the original one. Therefore, we conclude that the output sequence can be just the clean speech and does not need to align with the continuous input condition sequence. 
Following the approach in SpeechX \cite{wang2024speechx}, we conduct experiments where the input sequences are discrete token embeddings rather than continuous log-mel spectrograms, referred to as \textit{Discrete IO}. For both the input reference speech and the mixture speech, instead of using continuous log-mel spectrograms, we utilize the first two layers of the RVQ output of the neural audio codec. We then use two learnable embedding matrices to embed the discrete tokens and summarize these embeddings into a single embedding. This single embedding is then fed into the AR model. We observe that these discretized representations perform worse than the continuous approach. One possible reason for this could be that the neural audio codec's discrete representation may not be optimal for our TSE task with a small-scale decoder-only LM. Additionally, since the audio codec is typically trained on clean speech, it might miss crucial information from the mixture. A potential future work could involve developing audio codecs that can effectively handle mixture speech.

We also utilize WavLM features instead of the neural audio codec as the embedding features. We use the output from the 6th hidden layer of WavLM as input features for both the reference and the mixture. Additionally, we apply the concatenation technique used in TSELM \cite{tang2024tselm} for mixture representations. Following the approach in SELM \cite{selm}, the output of the AR model is the Kmeans discrete representations of the target speech. After obtaining the target discrete embeddings, we use a conformer detokenizer to reconstruct the continuous embeddings, as done in SELM \cite{selm}. Similar as \cite{tang2024tselm}, this approach, referred to as \textit{WavLM input}, results in poor speaker similarity, likely due to the discretization process that loses speaker information. 




\section{Conclusion}

We propose LauraTSE, an Auto-Regressive (AR) Decoder-Only Language Model (LM) designed for Target Speaker Extraction. It consists of a small-scale AR decoder-only LM that predicts the coarse-grained information of the target speech using the continuous representations of the reference and the mixed speech, and a one-step encoder-only LM that captures the fine-grained acoustic details. Extensive experiments demonstrate the model's capability in promising speech quality, intelligibility, and speaker similarity.



\bibliographystyle{IEEEtran}
\bibliography{paper}

\begin{thebibliography}{10}
\providecommand{\url}[1]{#1}
\csname url@samestyle\endcsname
\providecommand{\newblock}{\relax}
\providecommand{\bibinfo}[2]{#2}
\providecommand{\BIBentrySTDinterwordspacing}{\spaceskip=0pt\relax}
\providecommand{\BIBentryALTinterwordstretchfactor}{4}
\providecommand{\BIBentryALTinterwordspacing}{\spaceskip=\fontdimen2\font plus
\BIBentryALTinterwordstretchfactor\fontdimen3\font minus \fontdimen4\font\relax}
\providecommand{\BIBforeignlanguage}[2]{{%
\expandafter\ifx\csname l@#1\endcsname\relax
\typeout{** WARNING: IEEEtran.bst: No hyphenation pattern has been}%
\typeout{** loaded for the language `#1'. Using the pattern for}%
\typeout{** the default language instead.}%
\else
\language=\csname l@#1\endcsname
\fi
#2}}
\providecommand{\BIBdecl}{\relax}
\BIBdecl

\bibitem{zmolikova2023neural}
K.~Zmolikova, M.~Delcroix, T.~Ochiai, K.~Kinoshita, J.~{\v{C}}ernock{\`y}, and D.~Yu, ``Neural target speech extraction: An overview,'' \emph{IEEE Signal Processing Magazine}, vol.~40, no.~3, pp. 8--29, 2023.

\bibitem{luo2019conv}
Y.~Luo and N.~Mesgarani, ``Conv-tasnet: Surpassing ideal time--frequency magnitude masking for speech separation,'' \emph{IEEE/ACM transactions on audio, speech, and language processing}, vol.~27, no.~8, pp. 1256--1266, 2019.

\bibitem{spex_plus}
M.~Ge, C.~Xu, L.~Wang, E.~S. Chng, J.~Dang, and H.~Li, ``Spex+: A complete time domain speaker extraction network,'' in \emph{Proceeding of Interspeech}, 2020, pp. 1406--1410.

\bibitem{sepformer}
C.~Subakan, M.~Ravanelli, S.~Cornell, M.~Bronzi, and J.~Zhong, ``Attention is all you need in speech separation,'' in \emph{Proceeding of IEEE International Conference on Acoustics, Speech and Signal Processing (ICASSP)}, 2021, pp. 21--25.

\bibitem{sef_net}
B.~Zeng, H.~Suo, Y.~Wan, and M.~Li, ``Sef-net: Speaker embedding free target speaker extraction network,'' in \emph{Proceeding of Interspeech}, 2023, pp. 3452--3456.

\bibitem{distortion}
P.~Wang, K.~Tan \emph{et~al.}, ``Bridging the gap between monaural speech enhancement and recognition with distortion-independent acoustic modeling,'' \emph{IEEE/ACM Transactions on Audio, Speech, and Language Processing}, vol.~28, pp. 39--48, 2019.

\bibitem{fang2021variational}
H.~Fang, G.~Carbajal, S.~Wermter, and T.~Gerkmann, ``Variational autoencoder for speech enhancement with a noise-aware encoder,'' in \emph{Proceeding of IEEE International Conference on Acoustics, Speech and Signal Processing (ICASSP)}.\hskip 1em plus 0.5em minus 0.4em\relax IEEE, 2021, pp. 676--680.

\bibitem{bie2022unsupervised}
X.~Bie, S.~Leglaive, X.~Alameda-Pineda, and L.~Girin, ``Unsupervised speech enhancement using dynamical variational autoencoders,'' \emph{IEEE/ACM Transactions on Audio, Speech, and Language Processing}, vol.~30, pp. 2993--3007, 2022.

\bibitem{lu2022conditional}
Y.-J. Lu, Z.-Q. Wang, S.~Watanabe, A.~Richard, C.~Yu, and Y.~Tsao, ``Conditional diffusion probabilistic model for speech enhancement,'' in \emph{Proceeding of IEEE International Conference on Acoustics, Speech and Signal Processing (ICASSP)}.\hskip 1em plus 0.5em minus 0.4em\relax Ieee, 2022, pp. 7402--7406.

\bibitem{target_diff}
N.~Kamo, M.~Delcroix, and T.~Nakatani, ``Target speech extraction with conditional diffusion model,'' in \emph{Proceeding of Interspeech}, 2023, pp. 176--180.

\bibitem{tokensplit}
H.~Erdogan, S.~Wisdom, X.~Chang, Z.~Borsos, M.~Tagliasacchi, N.~Zeghidour, and J.~R. Hershey, ``Tokensplit: Using discrete speech representations for direct, refined, and transcript-conditioned speech separation and recognition,'' in \emph{Proceeding of Interspeech}, 2023, pp. 3462--3466.

\bibitem{zhang2025anyenhance}
J.~Zhang, J.~Yang, Z.~Fang, Y.~Wang, Z.~Zhang, Z.~Wang, F.~Fan, and Z.~Wu, ``Anyenhance: A unified generative model with prompt-guidance and self-critic for voice enhancement,'' \emph{arXiv preprint arXiv:2501.15417}, 2025.

\bibitem{kang2025llase}
B.~Kang, X.~Zhu, Z.~Zhang, Z.~Ye, M.~Liu, Z.~Wang, Y.~Zhu, G.~Ma, J.~Chen, L.~Xiao \emph{et~al.}, ``Llase-g1: Incentivizing generalization capability for llama-based speech enhancement,'' \emph{arXiv preprint arXiv:2503.00493}, 2025.

\bibitem{vae}
R.~Wang, L.~Li, and T.~Toda, ``Dual-channel target speaker extraction based on conditional variational autoencoder and directional information,'' \emph{IEEE/ACM Transactions on Audio, Speech, and Language Processing}, vol.~32, pp. 1968--1979, 2024.

\bibitem{tang2024tselm}
B.~Tang, B.~Zeng, and M.~Li, ``Tselm: Target speaker extraction using discrete tokens and language models,'' \emph{arXiv preprint arXiv:2409.07841}, 2024.

\bibitem{wavlm}
S.~Chen, C.~Wang, Z.~Chen, Y.~Wu, S.~Liu, Z.~Chen, J.~Li, N.~Kanda, T.~Yoshioka, X.~Xiao, J.~Wu, L.~Zhou, S.~Ren, Y.~Qian, Y.~Qian, J.~Wu, M.~Zeng, X.~Yu, and F.~Wei, ``Wavlm: Large-scale self-supervised pre-training for full stack speech processing,'' \emph{IEEE Journal of Selected Topics in Signal Processing}, vol.~16, no.~6, pp. 1505--1518, 2022.

\bibitem{wang2024speechx}
X.~Wang, M.~Thakker, Z.~Chen, N.~Kanda, S.~E. Eskimez, S.~Chen, M.~Tang, S.~Liu, J.~Li, and T.~Yoshioka, ``Speechx: Neural codec language model as a versatile speech transformer,'' \emph{IEEE/ACM Transactions on Audio, Speech, and Language Processing}, 2024.

\bibitem{du2023lauragpt}
Z.~Du, J.~Wang, Q.~Chen, Y.~Chu, Z.~Gao, Z.~Li, K.~Hu, X.~Zhou, J.~Xu, Z.~Ma \emph{et~al.}, ``Lauragpt: Listen, attend, understand, and regenerate audio with gpt,'' \emph{arXiv preprint arXiv:2310.04673}, 2023.

\bibitem{selm}
Z.~Wang, X.~Zhu, Z.~Zhang, Y.~Lv, N.~Jiang, G.~Zhao, and L.~Xie, ``Selm: Speech enhancement using discrete tokens and language models,'' in \emph{Proceeding of IEEE International Conference on Acoustics, Speech and Signal Processing (ICASSP)}, 2024, pp. 11\,561--11\,565.

\bibitem{unit_hifi}
P.~Mousavi, J.~Duret, S.~Zaiem, L.~{Della Libera}, A.~Ploujnikov, C.~Subakan, and M.~Ravanelli, ``How should we extract discrete audio tokens from self-supervised models?'' in \emph{Proceeding of Interspeech}, 2024, pp. 2554--2558.

\bibitem{dasb}
P.~Mousavi, L.~Della~Libera, J.~Duret, A.~Ploujnikov, C.~Subakan, and M.~Ravanelli, ``Dasb--discrete audio and speech benchmark,'' \emph{arXiv preprint arXiv:2406.14294}, 2024.

\bibitem{du2024funcodec}
Z.~Du, S.~Zhang, K.~Hu, and S.~Zheng, ``Funcodec: A fundamental, reproducible and integrable open-source toolkit for neural speech codec,'' in \emph{Proceeding of IEEE International Conference on Acoustics, Speech and Signal Processing (ICASSP)}, 2024, pp. 591--595.

\bibitem{gulati2020conformer}
A.~Gulati, J.~Qin, C.-C. Chiu, N.~Parmar, Y.~Zhang, J.~Yu, W.~Han, S.~Wang, Z.~Zhang, Y.~Wu, and R.~Pang, ``Conformer: Convolution-augmented transformer for speech recognition,'' in \emph{Proceeding of Interspeech}, 2020, pp. 5036--5040.

\bibitem{wang2024wesep}
S.~Wang, K.~Zhang, S.~Lin, J.~Li, X.~Wang, M.~Ge, J.~Yu, Y.~Qian, and H.~Li, ``Wesep: A scalable and flexible toolkit towards generalizable target speaker extraction,'' in \emph{Proceeding of Interspeech}, 2024, pp. 4273--4277.

\bibitem{usef_tse}
B.~Zeng and M.~Li, ``Usef-tse: Universal speaker embedding free target speaker extraction,'' \emph{IEEE Transactions on Audio, Speech and Language Processing}, 2025.

\bibitem{sato2024speakerbeam}
H.~Sato, T.~Moriya, M.~Mimura, S.~Horiguchi, T.~Ochiai, T.~Ashihara, A.~Ando, K.~Shinayama, and M.~Delcroix, ``Speakerbeam-ss: Real-time target speaker extraction with lightweight conv-tasnet and state space modeling,'' in \emph{Proceeding of Interspeech}, 2024, pp. 5033--5037.

\bibitem{tsunoo2024decoder}
E.~Tsunoo, H.~Futami, Y.~Kashiwagi, S.~Arora, and S.~Watanabe, ``Decoder-only architecture for streaming end-to-end speech recognition,'' in \emph{Proceeding of Interspeech}, 2024, pp. 4463--4467.

\bibitem{librispeech}
V.~Panayotov, G.~Chen, D.~Povey, and S.~Khudanpur, ``Librispeech: an asr corpus based on public domain audio books,'' in \emph{Proceeding of IEEE International Conference on Acoustics, Speech and Signal Processing (ICASSP)}, 2015, pp. 5206--5210.

\bibitem{librimix}
J.~Cosentino, M.~Pariente, S.~Cornell, A.~Deleforge, and E.~Vincent, ``Librimix: An open-source dataset for generalizable speech separation,'' \emph{arXiv preprint arXiv:2005.11262}, 2020.

\bibitem{rix2001perceptual}
A.~W. Rix, J.~G. Beerends, M.~P. Hollier, and A.~P. Hekstra, ``Perceptual evaluation of speech quality (pesq)-a new method for speech quality assessment of telephone networks and codecs,'' in \emph{Proceeding of IEEE International Conference on Acoustics, Speech and Signal Processing (ICASSP)}, vol.~2, 2001, pp. 749--752.

\bibitem{taal2010short}
C.~H. Taal, R.~C. Hendriks, R.~Heusdens, and J.~Jensen, ``A short-time objective intelligibility measure for time-frequency weighted noisy speech,'' in \emph{Proceeding of IEEE International Conference on Acoustics, Speech and Signal Processing (ICASSP)}, 2010, pp. 4214--4217.

\bibitem{dnsmos}
C.~K. Reddy, V.~Gopal, and R.~Cutler, ``Dnsmos p. 835: A non-intrusive perceptual objective speech quality metric to evaluate noise suppressors,'' in \emph{Proceeding of IEEE International Conference on Acoustics, Speech and Signal Processing (ICASSP)}, 2022, pp. 886--890.

\bibitem{mittag2021nisqa}
G.~Mittag, B.~Naderi, A.~Chehadi, and S.~Möller, ``Nisqa: A deep cnn-self-attention model for multidimensional speech quality prediction with crowdsourced datasets,'' in \emph{Proceeding of Interspeech}, 2021, pp. 2127--2131.

\bibitem{saeki2024speechbertscore}
T.~Saeki, S.~Maiti, S.~Takamichi, S.~Watanabe, and H.~Saruwatari, ``Speechbertscore: Reference-aware automatic evaluation of speech generation leveraging nlp evaluation metrics,'' in \emph{Proceeding of Interspeech}, 2024, pp. 4943--4947.

\bibitem{hubert}
W.-N. Hsu, B.~Bolte, Y.-H.~H. Tsai, K.~Lakhotia, R.~Salakhutdinov, and A.~Mohamed, ``Hubert: Self-supervised speech representation learning by masked prediction of hidden units,'' \emph{IEEE/ACM transactions on audio, speech, and language processing}, vol.~29, pp. 3451--3460, 2021.

\bibitem{dwer}
Z.-Q. Wang, H.~Erdogan, S.~Wisdom, K.~Wilson, D.~Raj, S.~Watanabe, Z.~Chen, and J.~R. Hershey, ``Sequential multi-frame neural beamforming for speech separation and enhancement,'' in \emph{Proceeding of IEEE Spoken Language Technology Workshop (SLT)}, 2021, pp. 905--911.

\bibitem{whisper}
A.~Radford, J.~W. Kim, T.~Xu, G.~Brockman, C.~McLeavey, and I.~Sutskever, ``Robust speech recognition via large-scale weak supervision,'' in \emph{Proceeding of International conference on machine learning}, 2023, pp. 28\,492--28\,518.

\bibitem{wespeaker}
H.~Wang, C.~Liang, S.~Wang, Z.~Chen, B.~Zhang, X.~Xiang, Y.~Deng, and Y.~Qian, ``Wespeaker: A research and production oriented speaker embedding learning toolkit,'' in \emph{Proceeding of IEEE International Conference on Acoustics, Speech and Signal Processing (ICASSP)}, 2023, pp. 1--5.

\bibitem{voxceleb}
A.~Nagrani, J.~S. Chung, W.~Xie, and A.~Zisserman, ``Voxceleb: Large-scale speaker verification in the wild,'' \emph{Comput. Speech Lang.}, vol.~60, no.~C, 2020.

\bibitem{data_scale_prove}
T.~Brown, B.~Mann, N.~Ryder, M.~Subbiah, J.~D. Kaplan, P.~Dhariwal, A.~Neelakantan, P.~Shyam, G.~Sastry, A.~Askell \emph{et~al.}, ``Language models are few-shot learners,'' \emph{Advances in neural information processing systems}, vol.~33, pp. 1877--1901, 2020.

\bibitem{ste}
Y.~Bengio, N.~L{\'{e}}onard, and A.~C. Courville, ``Estimating or propagating gradients through stochastic neurons for conditional computation,'' \emph{CoRR}, vol. abs/1308.3432, 2013.

\end{thebibliography}

\end{document}